\documentclass{article}

\usepackage[nonatbib,preprint]{neurips_2020}
\usepackage[utf8]{inputenc} 
\usepackage[T1]{fontenc}    
\usepackage{hyperref}       
\usepackage{url}            
\usepackage{booktabs}       
\usepackage{amsfonts}       
\usepackage{nicefrac}       
\usepackage{microtype}      
\usepackage{amsmath,amsfonts,amssymb}
\usepackage[ruled,longend]{algorithm2e}
\SetKw{KwBy}{by}
\usepackage{float}
\usepackage{multicol}
\usepackage{microtype}
\usepackage{graphicx}
\usepackage{subfigure}
\usepackage{booktabs}
\usepackage[tableposition = top]{caption}
\usepackage{siunitx}
\usepackage{tabularx}

\usepackage[square,numbers]{natbib} 
\title{Generalizing Gain Penalization for Feature Selection in Tree-based Models}

\author{%
  Bruna Wundervald \thanks{https://brunaw.com/} \\
  Hamilton Institute, \\
  Maynooth University\\
  \texttt{bruna.wundervald@mu.ie} \\ \And
    Andrew Parnell \\ Hamilton Institute, \\ Maynooth University\\
  \texttt{andrew.parnell@mu.ie} \\
  \And Katarina Domijan \\ Hamilton Institute, \\ Maynooth University\\
  \texttt{katarina.domijan@mu.ie} 
  }

\begin{document}

\maketitle

\begin{abstract}
We develop a new approach for feature selection via gain penalization in tree-based models. First, we show that previous methods do not perform sufficient regularization and often exhibit sub-optimal out-of-sample performance, especially when correlated features are present. Instead, we develop a new gain penalization idea that exhibits a general local-global regularization for tree-based models. The new method allows for more flexibility in the choice of feature-specific importance weights. We validate our method on both simulated and real data and implement it as an extension of the popular \texttt{R} package \texttt{ranger}.
\end{abstract}

\section{Introduction}

In many Machine Learning problems, features can be hard or economically 
expensive to obtain, and some may be irrelevant or poorly linked to the target. 
For these reasons, reducing the number of features is an important
task when building a model, and benefits the data visualization and model performance,  
whilst reducing storage and training time requirements \citep{Guyon}. However, for tree-based methods, there is no standard procedure for
feature selection or regularization in the literature, as one would find 
for Linear Regression and the LASSO \citep{Tibshirani1991} for example. 
Performing feature selection in trees can be difficult, as they struggle 
to detect highly correlated features and their 
feature importance measures are not fully trustworthy \citep{phdthesisRF}. Several methods to tackle this problem have been recently proposed, including \citet{DiazUriarte2007}, \citet{reg_tree_fried}, and \citet{rrf_paper}. 

In \citet{reg_tree_fried}, the authors treat trees as 
parametric models and use procedures analogous to LASSO-type shrinkage
methods, by penalizing the coefficients of the 
base learners and reducing the redundancy in each path from the root 
node to a leaf node. However, their selected features can
still be redundant, since the focus is on reducing the number of rules 
instead of the number of features. \citet{DiazUriarte2007} focuses on gene selection for classification. 
The authors propose an iterative tool that eliminates the least 
important features (in fractions of the number of features, $p$) and updates the algorithm at 
each iteration. The complication is that the method will
always be either computationally expensive, if $p$ is low, 
or will eliminate too many features at once, which can exclude useful 
or interaction features. Besides, the method does not generalize to other
dataset contexts or tasks, such as regression. In the contrasting approach of \citet{rrf_paper, guided}, the authors regularize Random Forests by gain penalization. Their method consists of letting the features only be picked by a Random Forest if their penalized 
(weighted) gain is still high. They make 
recommendations on how to set the penalization coefficients and 
present their implementation in the \texttt{RRF} package for \texttt{R} \citep{rmanual}. However, the authors give no further guidelines on 
how to generalize their method for other models and
penalization types and do not explore the influence of hyperparameters 
on the algorithm.

We develop a gain penalization approach that is more general and widely applicable than those mentioned above. In particular, our contributions are:

\begin{itemize}
\setlength\itemsep{0.1em}
  \item We provide a richer means by which local and global regularization of features can be balanced and allow for bespoke local regularization functions for domain-specific applications. 
  \item We generalize gain penalization
  to multiple tree-based methods (CART, Bagging, Random Forests), for both regression and classification. 
  \item We propose different techniques for setting the regularization parameters
  and discuss how they affect the final results, with real and simulated
  examples. 
  \item We make available a faster implementation of the regularization method, included in the very widely used \texttt{ranger} package. 
\end{itemize}

The paper is structured as follows. Section 2 explains the problem setup, followed by the generalization of gain penalization
in Section 3. In Section 4, we present the results 
for simulated and real data. Section 5 explains the implementation
details, and Section 6 has the conclusions and future work. 

\section{Problem setup} 

Consider a set of training target-feature pairs 
$(Y_i, \mathbf{x}_i) \in \mathbb{R} \times \mathbb{R}^{p}$, 
with $i = 1, \dots, N$ indexing the observations with $p$ being 
the total number of features. In general, we can estimate an $\hat f$ that
describes how the features $\mathbf{x}_i$ relate to $Y_i$ and use
it for prediction or inference. However, 
not all features need to be involved in $\hat f$. Especially for 
tree-based models, the occurrence of noisy or correlated features
can badly influence the results \citep{cond_var_imp}. 
Our interest here relies on estimating
$\hat f$ such that it will only use the matrix $\mathbf{x}_{A}$, composed by the sub-vectors of $\mathbf{x} \in \mathbb{R}^{p}$
indexed by $A$,  $A \subset \{1,\dots p \}$, which
should contain the optimal set of features (it produces similar or equal prediction errors as the full matrix of features), that is 
potentially of a much smaller dimension.

\subsection{Trees}



Trees are a particular case of
non-linear models, that recursively partition the feature space, 
resulting in a local model for each estimated region 
\citep{breiman1984CART}. They learn the features directly from
the training data, creating an adaptive basis function model (ABM)
\citep{Murphy2012MachineL} of the form

\begin{equation}
f(\mathbf{x}) = \mathbb{E}[y \mid \mathbf{x}] = 
\sum_{r = 1}^{R} w_r \mathbb{I}(\mathbf{x} \in R_r) = 
\sum_{r = 1}^{R} w_r \phi_r(\mathbf{x}; \mathbf{v}_r),
\end{equation}

where $R_r$ is the $r$-th region, $w_r$ is the prediction given to
this region and $\mathbf{v}_r$ represents the splitting feature chosen to 
and the correspondent splitting value. These algorithms are fitted 
using a greedy procedure, that computes a locally optimal maximum 
likelihood estimator by finding the splits that lead to the
minimization of a cost function. 
For regression, the cost function of a decision $\mathbb{D}$ is frequently defined as $cost(\mathbb{D}) = \sum_{i \in \mathbb{D}} (y_i - \bar{y})^{2}$, 
where $\bar{y} = (\sum_{i \in \mathbb{D}} y_i) |\mathbb{D}|^{-1}$
is the mean of the observations in the specified region,
while for classification this function is replaced by the
misclassification rate, or $cost(\mathbb{D}) = |\mathbb{D}|^{-1} \sum_{i \in \mathbb{D}} \mathbb{I}(y_i \neq \hat y)$. The gain of a new split
is a normalized measure of the cost reduction, given by

\begin{equation}
\begin{split}
\Delta(i, t) = & cost(\mathbb{D}) - \left(
\frac{|\mathbb{D}_{LN_{(i, t)}}|}{|\mathbb{D}|} cost (\mathbb{D}_{LN_{(i, t)}}) +
\frac{|\mathbb{D}_{RN_{(i, t)}}|}{|\mathbb{D}|} cost (\mathbb{D}_{RN_{(i, t)}})
\right ), 
\label{eq:cost_tree}
\end{split}
\end{equation}

for feature $i$ at splitting point $t$, while $\mathbb{D}$ is 
related to the previous estimated split,
$LN=\text{(left candidate node)}$ and $RN=\text{(right candidate node)}$. 
The global importance value is given by accumulating the
gain over a feature, 
$\Delta(i) = \sum_{t \in \mathbb{S}_i} \Delta(i, t)$, 
where $\mathbb{S}_i$ now represents all the splitting points 
used in a tree for the $i-$th feature.

\subsection{Ensembles}

Trees are known to be high variance estimators: 
small changes in the data can lead 
to the estimation of a completely different tree 
\citep{Murphy2012MachineL}. One way to increase stability is to use the property 
that an average of many estimates has a smaller variance than one estimate, 
and grow many trees from re-samples of the data. Averaging such results give us a bagged ensemble \citep{Breiman1996} of the form $
\hat f(\mathbf{x}) = \sum_{n = 1}^{N_{tree}} \frac{1}{N_{tree}} \hat f_n(\mathbf{x}),$ where $\hat f_n$ corresponds to the $n$-th tree. The Random Forest \citep{Breiman2001}
algorithm is defined by allowing only a random 
subset $m$ of the features to be the candidates in each split.
For ensembles, the importance value for a feature gets averaged over all the trees, or

\begin{equation}
Imp_{i} = \frac{1}{N_{tree}} \sum_{n = 1}^{N_{tree}} \Delta(i)_{n},
\end{equation}

for feature $i$. Moreover, the performance of the
trees in a Random Forest relies on the number of features
tried at each split, called $\texttt{mtry}$ here, 
as when $\texttt{mtry} \rightarrow 1$, the 
larger the variance of each tree, but the more effective will 
be the averaging process, and vice versa
\citep{phdthesisRF}.

\subsection{Regularization by gain penalization}

In \citet{rrf_paper}, the authors first discuss the regularization of
Random Forests by gain penalization. The
\textit{Regularized Random Forest} (\textit{RRF}) 
proposes weighting the gains of the splits during the greedy 
procedure, guiding the feature choosing of the model.
The regularized gain is defined as

\begin{equation}
\text{Gain}_{R}(\mathbf{X}_{i}, t) = 
\begin{cases}
\lambda_{i} \Delta(i, t), \thinspace  i \notin \mathbb{U} \text{ and} \\
\Delta(i, t), \thinspace  i \in \mathbb{U}, 
\end{cases}
\label{eq:grrf}
\end{equation}

where $\mathbb{U}$ is the set of indices of the features previously 
used, $\mathbf{X}_{i}$ is the candidate feature, and
$t$ the candidate splitting point. The $\lambda_i \in (0, 1]$ 
parameter is the penalty coefficient that
controls the amount of regularization each feature receives. 
A feature  is penalized if it is new to 
the whole ensemble, as the method has a memory of 
which features were already used. Naturally, $\lambda_i$ can be a constant value for all the features but 
ideally, there should be a regularization parameter for each feature 
that best represents the information they carry about the target. In 
\citet{guided}, the authors develop this idea by introducing the 
\textit{Guided RRF}. It consists of first running a Standard Random
Forest (\texttt{mtry} $\approx \sqrt{p}$, number of trees = 500), producing an importance measure for each feature and scaling this measure, in order to find $Imp_{i}^{'}  = \frac{Imp_i}{max_{j = 1}^{P} Imp_j}$,
where $Imp_i$ is the importance measure calculated for the
$i$-th feature in the Random Forest. The estimated normalized variable importance measures are considered jointly with a regularization parameter to create the overall gain penalization.



\section{Generalizing Gain Penalization} 

One of our goals with this work is to show how Equation 
\ref{eq:grrf} can be fully generalized in two
senses: in the algorithm type and the penalization coefficients. For the
algorithm, this means that the regularization method can be applied to 
any tree-based model  (single trees such as CART or ensembles).
As for the penalization coefficients, we generalize the method by 
proposing that $\lambda_i$ can be written as

\begin{equation}
\lambda_i = (1 - \gamma) \lambda_0 + \gamma g(\mathbf{x}_i), 
\label{eq:generalization}
\end{equation}

where $\lambda_0 \in [0, 1)$ is interpreted as the 
baseline regularization,  $g(\mathbf{x}_i)$ 
is a function of the  $i$-th feature, 
and $\gamma \in [0, 1)$ is their mixture parameter, 
with $\lambda_i \in [0, 1)$. The 
equation balances how much all features
should be jointly, or globally, penalized and how much will it be due 
to a local $g(\mathbf{x}_i)$. When $\gamma = 0$, we return
to what was proposed in \citet{rrf_paper}, and for $\gamma = 1$, the regularization is fully controlled by $g(\mathbf{x}_i)$. The $g(\mathbf{x}_i)$ should represent relevant information about the 
features, based on some characteristic of interest. 
It can include, for example, external information about the relationship between $\mathbf{x}_i$ and $\mathbf{y}$, thus this has inspiration on the
use of priors made in Bayesian methods. In the same fashion, the data will tell us how strong our assumptions about the penalization are, since even if we try to penalize a truly important feature, its gain will be high enough to overcome the penalization and the feature will get picked.

\subsection{Choosing $g(\mathbf{x}_i)$}


\textbf{Correlation}: A natural option for continuous features is just to 
use $g(\mathbf{x}_i)$  as the absolute value of the marginal 
correlation between $\mathbf{x}_i$ and $\mathbf{y}$, assuming a 
continuous target problem. It can be Pearson's,
Kendall's, Spearman's, or other correlation coefficient of preference
(the first is more suitable for ordinary numeric inputs, 
while the others will be more convenient for ordered
inputs \citep{chen2002correlation}). We can define it as
$g(\mathbf{x}_i) = |corr(\mathbf{y}, \mathbf{x}_i)|$.


\textbf{Entropy and Mutual Information}: 
A different situation is when the features are discrete. 
In information theory, 
Shannon's entropy \citep{shannon1948mathematical} is
a measure of the uncertainty of a (discrete) random feature. In short,
if a discrete feature $X$ has $K$ states, its entropy will
be calculated as $\mathbb{H}(X) = - \sum_{k = 1}^{K} p(X = k) \log_{2} p(X = k),\label{eq:entropy}$ where $\mathbb{H}(X) \in [0, \infty]$.  Higher entropy will mean more uncertainty, 
so it is reasonable to give more weight to features with 
lower uncertainties. One can use a normalized
version of the entropy calculated for each $\mathbf{x}_i$, or 
$g(\mathbf{x}_i) = 1 - \frac{\mathbb{H}(\mathbf{x}_{i})}
{max_{j=1}^{P} \mathbb{H}(\mathbf{x}_{j})}, $

compelling the features with lower entropy to have larger penalization 
coefficients. Similarly, we can also use the Mutual Information function, for which the similarity between a joint distribution $p(X, Y)$ and a factored distribution $p(X)p(Y)$ and we calculate
$\text{MutInf}(X; Y) = \sum_{x} \sum_{y} p(x, y) \log \frac{p(x, y)}{p(x)p(y)}
$ for two features $X$ and $Y$ \citep{Murphy2012MachineL}.
Recalling the Entropy equation, it is easy to see that the Mutual
Information value is the reduction in the uncertainty about $Y$ when we
observe $X$, so it can be straightforwardly used as $g(\mathbf{x}_i) = \frac{\text{MutInf}(\mathbf{x}_i, \mathbf{y})}
{max_{j=1}^{P}\text{MutInf}(\mathbf{x}_j, \mathbf{y})}$.

\textbf{Boosted}:  When there is no interest in differentiating 
continuous or
discrete features, one can use a \textit{Boosted}  $g(\mathbf{x}_i)$. 
Such functions depend on previously run 
machine learning models that provide an importance value for 
the features. The term \textit{Boosted} 
is to introduce some familiarity, since we can arguably see the 
algorithm as an heterogenous Boosting
\citep{nascimento2009ensembling}
applied to the features instead of the observations. Examples of
Machine Learning algorithms that allow for the calculation of
an importance value include: tree ensembles, Generalized Linear Models \citep{nelder1972generalized}, where
the normalized absolute parameter coefficients can be interpreted as importance values, 
and Support Vector Machines \citep{Hastie} that produce importance
values via sensitivity analysis \citep{CORTEZ20131, rminer}. 
We should note that each family of algorithms will have its specific characteristics and preferences towards the features, that might
need to be taken into account.

\textbf{Combination}: Another possibility is combining two or more
$g(\mathbf{x}_i)$.  Objectively speaking, 
some functions will be more appropriate to one type of feature than others. 
As an example, one could 
combine a \textit{Boosted}
method with the marginal correlations between the target and 
each feature, letting the absolute values of the correlations compose 
$g(\mathbf{x}_i)$ if the 
correlation is over a certain threshold $\epsilon$, and use
$Imp_{i}^{'}$ from a previously run algorithm otherwise.  



\textbf{Depth parameter}: Growing very bushy trees with new features is 
not desirable when we want to use a small set of features. 
Following \citet{Chipman2010}, where 
the authors use prior distributions 
for whether a new feature should be picked in a Bayesian Regression Tree, 
we introduce the idea of increasing the penalization given the
current depth of the tree. Their priors take into account the current 
depth of a tree, so when a tree is already deep the
priors get less concentrated in high probability regions, resulting
in lesser bushier trees. In our work, a similar idea is applied by setting 

\begin{equation}
Gain_{R}(\mathbf{X}_{i}, t, \mathbb{T}) = 
\begin{cases}
\lambda_{i}^{d_{\mathbb{T}}} \Delta(i, t),
\thinspace i \notin \mathbb{U} \text{ and} \\
\Delta(i, t), \thinspace i \in  \mathbb{U}, 
\end{cases}
\label{eq:grrf_depth}
\end{equation}

where $d_{\mathbb{T}}$ is the current depth of the $\mathbb{T}$ tree, 
$\mathbb{T} = (1, \dots, \texttt{ntree})$, for the $i$-th feature.  The aim here is to reduce the gains of the features
if they are to be picked in a deep node, preventing new
features to appear at the bottom of trees unless their gains
are exceptionally high.

\subsection{Issues \& Details} 

\textbf{Feature masking effect:} Tree-based models often suffer
from feature masking effects (\citet{phdthesisRF}). For example, 
in a tree, some feature $X_j$ might never occur in the 
algorithm if it leads to splits slightly worse than some other
feature $X_i$, so if  $X_i$ is removed, $X_j$ can 
prominently occur. This should be overcomed by ensembles like Random Forests, 
as selecting only $m$ features to pick from decorrelates the trees, but if we regularize the Random Forests, the problem remains.
If weak features end up being first picked by the model,
their gains will have an unfair advantage against
the other features, which will be penalized. This situation is
easily fixed with hyperparameter tuning for \texttt{mtry}.

\textbf{Correlated features:}
Random Forests are also biased
towards giving high importance to correlated 
features \citep{cond_var_imp}. Suppose
we have a subset $\mathbf{C} \subseteq \mathbf{X}$ of features
which are correlated. Ideally, we want to have only 
one or just a few of these features being selected to 
avoid redundancy, 
but Random Forests are not able to detect and eliminate correlated features. The regularized method automatically deals with the
correlated features, since when one of the features in $\mathbf{C}$
gets picked, the algorithm is less likely to pick the other correlated
features as well, given that a new feature needs to reduce the prediction error more drastically to be selected.

\section{Experiments}

This section shows experiments that evaluate the effects of 
different regularization types in simulated and real datasets using
the Random Forest algorithm. 

\subsection{Simulated data}

Consider now a set  $\mathbf{X} = (\mathbf{x}_{1},\dots, \mathbf{x}_{205})$ of features, all sampled from a Uniform[0, 1] distribution,
$n = 1000$. We generated a target of interest 
$\mathbf{Y} \in \mathbb{R}$  as 
 
\begin{equation}
\begin{split}
\mathbf{y} = & 0.8 sin(\mathbf{x}_1 \mathbf{x}_2) + 2 (\mathbf{x}_3
- 0.5)^2 + 1 \mathbf{x}_4 + 0.7 \mathbf{x}_5 + 
\sum_{j = 1}^{200} 0.9^{(j/3)} \mathbf{x}_{j+5} +  \sum_{j = 1}^{45} 0.9^{j} \mathbf{x}_5 + \mathbf{\epsilon}, 
\thinspace \mathbf{\epsilon} \sim N(0, 1), 
\end{split}
\end{equation}

inspired by the simulation equation proposed in \citet{Friedman1991}, 
totaling 250 features. This framework produces interesting relationships
between the target and the features: non-linearities 
($i = (1, 2, 3)$), decreasing importances ($i = (6, \dots, 205)$) and
correlations ($i = (5, 206, \dots, 250)$), inducing a more complicated 
scenario. We created 10 datasets, all randomly split into train and test set (80\%/20\%). For all the algorithms we fixed the number of trees at 100, varied
$\texttt{mtry} = (15, 45, 75, 105, 135, 165, 195, 225, 250)$ and our accuracy measure is the RMSE calculated in the test set. 
We used a standardized version of $\mathbf{y}$ and, 
in the following, 
the term \textit{number of selected features} represents
any feature with importance $\Delta(i) > 0$ in the final estimated model.

\subsection{Standard Random Forest}

As a benchmark, we run a Standard Random Forest for each of 
the 10 datasets and all the different values of \texttt{mtry}. The first \texttt{mtry} is what would be the default in a 
Standard RF, since $\sqrt{250} \approx 15$, and the last
is the total of features available. 
The resulting number of features used for all the models is always 
the maximum available (250) (Table \ref{table:std_rf}). If we consider the correlated features issue, this means that too many features are being picked, 
once we know that they become irrelevant in their joint presence. 
The $\text{RMSE}_{\text{test}}$ changes when \texttt{mtry} changes: 
when \texttt{mtry} = 45 is when we have the best results, 
meaning that the default
value ($\texttt{mtry} = \sqrt{p} \approx 15$) is not the best option.

\begin{table}[ht]
\centering
\caption{Standard Random Forest results.}

\label{table:std_rf}
\begin{tabular}{l  | c  c c c c c c c c c}

\toprule
   \texttt{mtry} &  15 & 45 & 75 & 105 & 135 & 165 & 195 & 225 & 250 \\ 
  \hline
  Features &  250 & 250 & 250 & 250 & 250  &  250  & 250 & 250 & 250 \\
  $\text{RMSE}_{\text{test}}$  & 0.51 & 0.46 & 0.47 & 0.47 & 0.49 & 0.49 & 0.48 &
  0.48  & 0.48 \\ 
   \bottomrule
\end{tabular}
\end{table}

\subsection{\textit{RRF} and \textit{GRRF}} 

The simplest version of the regularized algorithm happens when we 
set $\lambda_i$ to be a constant value (\textit{RRF} approach, \citet{grrf_paper}), having all 
features penalized by the same factor. We now
present the results when fitting this algorithm 
to the simulated data. We varied
$\lambda_i = \lambda_0 = (0.05, 0.12, 0.18, 0.25, 0.32, 0.39, 0.45, 0.52, 0.59, 0.65, 0.72,  0.79, 0.86, 0.92, 0.99)$, and tested all combinations between $\lambda_i$ and \texttt{mtry}. We also present the results of our Guided RRF (\textit{GRRF}), using $\gamma = (0.05, 0.12, 0.18, 0.25, 0.32, 0.39, 0.45, 0.52, 0.59, 0.65, 0.72,  0.79, 0.86, 0.92, 0.99)$ (recall Equation \ref{eq:grrf}). The models
were run using the  \texttt{RRF} \citep{grrf_paper} package for \texttt{R}
\citep{rmanual}. 

\begin{figure}
\centering
\includegraphics[width = 350pt, height = 160pt]{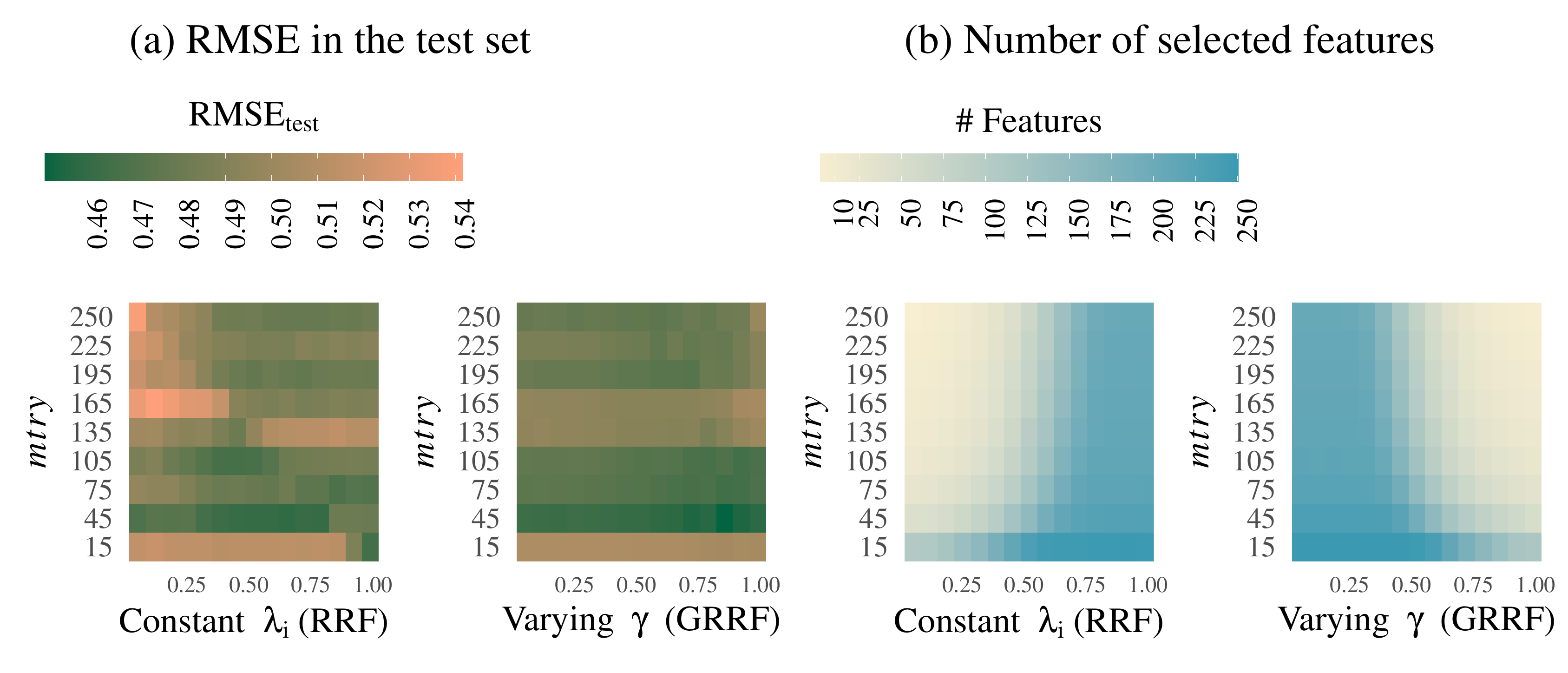}
\caption{(a) Tiles plot for the average of resulting $\text{RMSE}_{\text{test}}$ and (b) number of selected features in a \textit{Regularized Random Forest} and a \textit{Guided Regularized Random Forest} varying \texttt{mtry},  $\lambda_0$ and $\gamma$. The number of selected features has a different
relationship to \texttt{mtry} when comparing the two models. The $\text{RMSE}_\text{{test}}$ is generally lower for the \textit{GRRF}, and the regularization gets very affected by \texttt{mtry}.}
\label{fig:tile_rf} 
\end{figure}

Figure  \ref{fig:tile_rf} shows the results of the average $\text{RMSE}_{\text{test}}$ (left) and average number of features (right) in the 10 datasets for the two types of models. 
We can see a continuous transition on the number of features 
picked by the two models, but they present an inverse pattern regarding the
\texttt{mtry} and regularization parameters. For the \textit{RRF}, the lower the $\lambda_i$ and the 
higher the \texttt{mtry}, the less features are picked, but the $\text{RMSE}_{\text{test}}$ gets compromised. As for the \textit{GRRF}, the same happens but for the higher
$\gamma$ values. Though their number of features might be similar depending
on the hyperparameter values, the $\text{RMSE}_{\text{test}}$ values are in general lower
for the \textit{GRRF}, demonstrating how a more specific $\lambda_i$ 
for each feature improves the feature selection and the 
\textit{GRRF} has a clear advantage over the \textit{RRF}. 


\subsection{Generalized Regularization in Random Forests}

\begin{center}
\begin{figure}[hb]
\includegraphics[width=400pt,height=220pt]{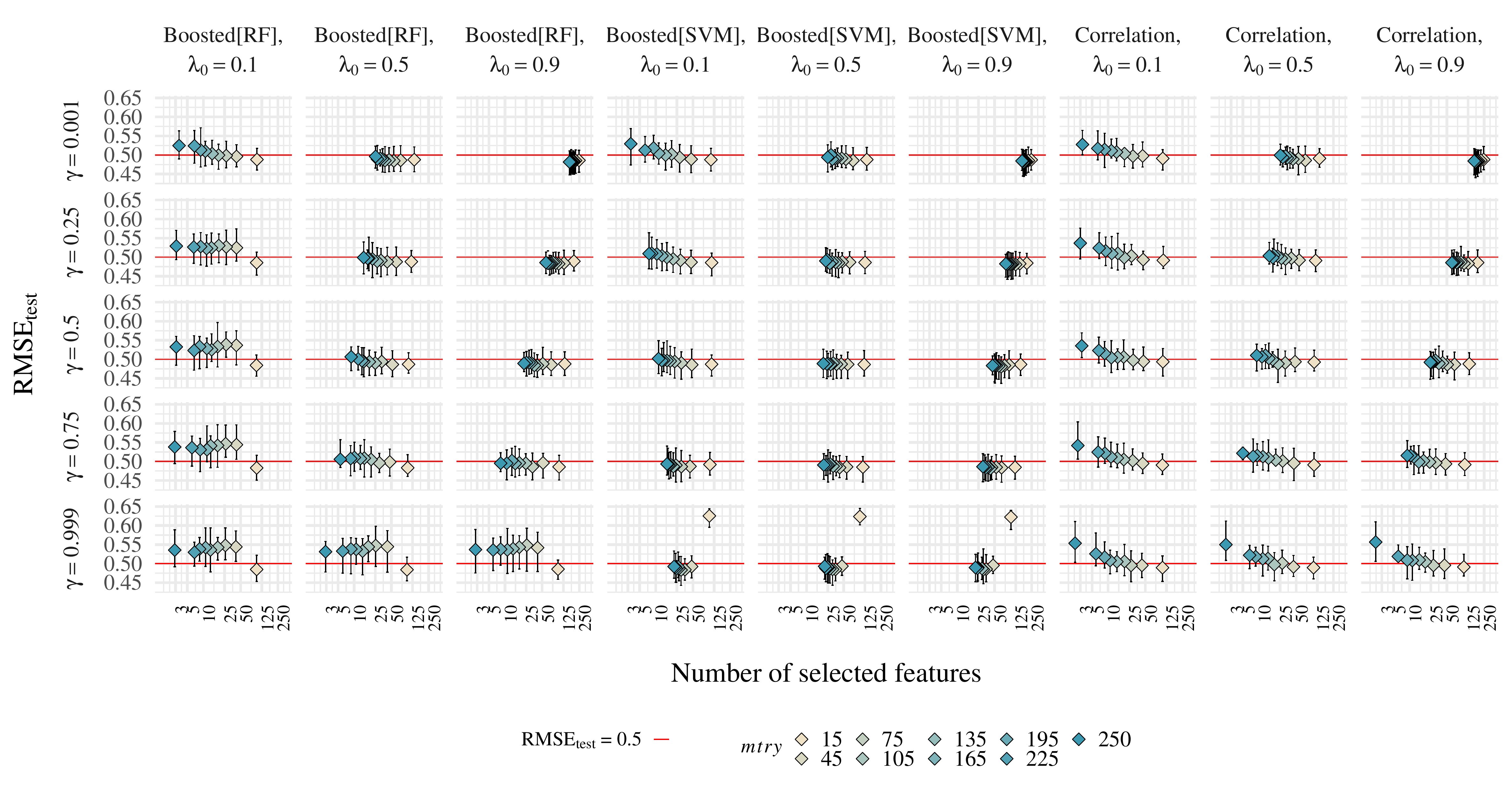}
\caption{$\text{RMSE}_{\text{test}}$ averages (with max-min intervals) and \textbf{log} number of features using the mixture of a $\lambda_{0}$ and a $g(\mathbf{x}_i)$, for  $g(\mathbf{x}_i) = (|corr(\mathbf{y}, \mathbf{x}_i)|,\thinspace \text{Boosted}_{RF}, \thinspace \text{Boosted}_{SVM}$). The x-axis shows the original scale, but the values are transformed to log. The models are use fewer features than the $\textit{GRRF}$ or Standard RF, with $\lambda_0$ and $\texttt{mtry}$ visibly affecting the results.}
\label{fig:mixtures}
\end{figure}
\end{center}

For this subsection, 
we vary $\lambda_0 = (0.1, 0.5, 0.9)$
and $\gamma = (0.001, 0.25, 0.5, 0.75, 0.99)$, use all 
combinations $(\gamma \times \lambda_0)$, first with $g(\mathbf{x}_i) = |corr(\mathbf{y}, \mathbf{x}_i)|$ and later using two
\textit{Boosted} methods, with a Standard Random Forest and with a 
Support Vector Machine. In Figure \ref{fig:mixtures} we can see that $\text{RMSE}_{\text{test}}$ values are either close 
or below
the 0.5 line. In comparison to Figure \ref{fig:tile_rf},
our algorithms are doing better, 
as we can spot many cases where the $\text{RMSE}_{\text{test}}$ is 
low while using very few 
features, especially when $g(\mathbf{x}_i) = \textit{Boosted}_{SVM}$. 
When $\gamma$ is low  the regularization is
primarily controlled by $\lambda_0$, and we spot a heavier influence of 
\texttt{mtry} on the number of selected features, which tends to decrease as $\lambda_0$ increases. When $\gamma$ is high the penalization values depend more on $g(\mathbf{x}_i)$, and the results vary less
regarding the values for $\lambda_0$ and \texttt{mtry}.

\begin{table}[h]
\centering
\caption{Percentages of the most important and of correlated features selected and $\text{RMSE}_{\text{test}}$, averaged by \texttt{mtry} and $\gamma$. When using $g(\mathbf{x}_i) = \text{Boosted}_{SVM}$,  we pick more of the important features, less of the correlated and have lower $\text{RMSE}_{\text{test}}$. The results for the GRRF are shown for comparison.}
\label{table:percentages}
\setlength\tabcolsep{2.15pt} 
\begin{tabular}{c | c  c  c || c  c  c || c  c  c }
\toprule 
  \multicolumn{1}{l}{} & \multicolumn{3}{c }{Correlation} & \multicolumn{3}{c}{$\text{Boosted}_{RF}$} & \multicolumn{3}{c}{$\text{Boosted}_{SVM}$} \\  
  \hline 
 $\lambda_0$ & \% Imp. & \% Corr. & $\text{RMSE}_{\text{test}}$ 
 & \% Imp. & \% Corr. & $\text{RMSE}_{\text{test}}$ 
 & \% Imp. & \% Corr. & $\text{RMSE}_{\text{test}}$ \\ 
  \hline
   0.10 & 65.2\% & 18.8\% & 0.51 &  69\% & 33.2\% & 0.52 
   & 71\%   & 21.6\% & 0.50 \\ 
  0.50 & 64.6\% & 19.0\% & 0.50  & 71.2\% & 28.2\% & 0.50
  & 69.8\% & 20.8\% & 0.49 
  \\ 
   0.90 & 64.6\% & 20.0\% & 0.49  & 67.6\% & 22.4\% & 0.48 &
   67.6\% & 22.4\% & 0.48 \\
   \hline
\end{tabular}

\begin{tabular}{c | c  c  c }
  \multicolumn{4}{c }{GRRF} \\  
  \hline 
 $\gamma$ & \% Imp. & \% Corr. & $\text{RMSE}_{\text{test}}$\\ 
  \hline
   $\approx$ 0.10 & 63.8\% & 33.1\% & 0.48 \\ 
  $\approx$ 0.50 &  67.6\% & 32.6\% & 0.48 \\
   $\approx$ 0.90 & 82.8\% & 32.0\% & 0.48 \\
 \bottomrule
\end{tabular}
\end{table}

A more in-depth analysis of  the results
can be seen in Table \ref{table:percentages}. We define 
the most informative features in the simulation as $\mathcal{V} = (\mathbf{x_1}, \mathbf{x_2}, \mathbf{x_3}, \mathbf{x_4}, \mathbf{x_5}, \underset{{i \in [6,205] \cap [0.9^{(i-5)/3}  > 0.01]}}{\mathbf{x_i}}).$ 
We do not include the last 45 features which are correlated and we ideally  want to avoid them.
We then calculate the percentage of important features from
the total of features that were picked, 
and from the correlated ones, which percentage of those 
was selected by the algorithm. 
So, for example, if an algorithm picked 10 features, 
3 of them being important, 5 being from the correlated group and 2
being "non-important", we calculate the proportion of important features as
$3/5$ and the proportion of correlated as $5/45$. With Table \ref{table:percentages}
we see that the proportion of important features is considerably higher 
for our approach with $g(\mathbf{x}_i) = \text{\textit{Boosted}}_{RF}$ and
$g(\mathbf{x}_i) = \text{\textit{Boosted}}_{SVM}$, and 
when we use  $g(\mathbf{x}_i) = |corr(\mathbf{y}, \mathbf{x}_i)|$
the algorithm picks less of the correlated variables. 
We also notice that the best results 
happened more or less when $\lambda_0 \leq 0.5$, when 
 $g(\mathbf{x}_i)$ has a higher influence in the penalization 
coefficients, so they are really helping the feature selection.
The GRRF algorithm usually picks more of the correlated
features and, once, more of the important ones,
but this model also picks more features in general.

\subsection{Real Data Classification}

  
  
   


\begin{table}[h]
\setlength\tabcolsep{1.6pt} 
\centering
\caption{Average \% of features used and average misclassification rates with standard deviation. The regularized models used far fewer features than a Standard RF. Clearly, our approaches uses far fewer features whilst maintaining competitive misclassification performance.}
\label{table:classification_results}

\begin{tabular}{l | c c c c || c c c c}
  \toprule
  \multicolumn{1}{l}{} & \multicolumn{4}{c }{\textbf{Percentage of features used}} & \multicolumn{4}{c}{\textbf{Misclassification rate in the test set}} \\  
  \hline
 Dataset & Std.RF & GRRF & $\text{Bst}_{RF}$ & Mut.Inf.  & Std.RF & GRRF & $\text{Bst}_{RF}$  &  Mut.Inf.   \\ 
  \hline
  \\[-8pt]

adenocarcinoma & 9.38 & 0.83 & 0.86 & 0.07 & 3.45 (0.0) & 11.03 (12) & \textbf{3.4} (0.0) & 10.77 (1.7) \\ 
brain & 25.06 & 1.44 & 1.46 & 0.14 & 8.00 (5.6) & 15.38 (14.4) & 15.00 (7) & \textbf{13.33 (10.5)} \\ 
breast 2 & 24.44 & 1.76 & 1.79 & 0.14 & 25.6 (5.4) & 20.7 (2) & \textbf{17.8 (7.2)} & 20.7 (7.7) \\ 
breast 3 & 38.58 & 1.95 & 2.00 & 0.28 & 27.6 (6.4) & 30.6 (4) & \textbf{28.1 (3.8)} & 29.3 (1.6) \\ 
colon & 34.44 & 2.53 & 2.60 & 0.44 & 4.76 (0.0) & \textbf{5.71 (2.1)} & 6.67 (0.0) & 7.78 (3) \\ 
leukemia & 8.07 & 1.26 & 1.25 & 0.05 & 0.0 (0.0) & \textbf{0.0 (0.0)} & \textbf{0.0 (0.0)} & \textbf{0.0 (0.0)} \\ 
lymphoma & 13.88 & 1.07 & 1.14 & 0.08 & 0.0 (0.0) & \textbf{0.0 (0.0)} & \textbf{0.0 (0.0)} & \textbf{0.0 (0.0)} \\ 
nci & 44.77 & 1.81 & 1.88 & 0.16 & 30.77 (9.4) & 38.95 (8) & \textbf{37.5 (7.6)} & 43.75 (0.0) \\ 
  prostate & 18.22 & 1.46 & 1.40 & 0.09 & \textbf{0.54 (1.2)} & \textbf{0.54 (1.2)} & 1.08 (1.5) & \textbf{0.54 (1.2)} \\ 
  srbct & 29.69 & 2.26 & 2.25 & 0.3 & 0.0 (0.0) & \textbf{0.0 (0.0)} & 0.91 (2) & 1.74 (3.9) \\ 
\bottomrule
\end{tabular}
\end{table}


Our experiments with real data consider gene classification micro-array 
datasets ([\citenum{ramaswamy2003molecular}], [\citenum{pomeroy2002prediction}],
[\citenum{van2002gene}], [\citenum{alon1999broad}], [\citenum{golub1999molecular}], [\citenum{alizadeh2000distinct}], 
[\citenum{ross2000systematic}], [\citenum{singh2002gene}], [\citenum{khan2001classification}]), with an average of 4787 features, 67 observations and 3 classes (details are in the Appendix A 1.1). Those are classical examples of 
"large p, small n" datasets, but our method generalizes to data from 
any contexts or sizes. As the goal here is to find the
best features to predict the gene classes, the experiment conducted for this section is different. We run the regularized models and
extract their selected features, that are later used in a Standard Random Forest, with which the misclassification 
rates are calculated. This is to mimic how such an approach can be used in practice, where first a discovery experiment is run to identify important features, then a subsequent algorithm is run on a new data set using all of the features. We set $\gamma = \lambda_0 = 0.5$, attributing
the same weight to the baseline regularization and to
$g(\mathbf{x}_i)$. We vary $\texttt{mtry} = (\sqrt{p}, 0.15p, 0.40p, 0.75p,  0.95p)$ and
$g(\mathbf{x}_i) = \Big(\text{Boosted}_{RF}, \thinspace \frac{\text{MutInf}(\mathbf{y}, \mathbf{x}_i)}{max_{j=1}^{P}\text{MutInf}( \mathbf{y}, \mathbf{x}_j)} \Big)$. For comparison, we also run a Standard RF 
and a \textit{GRRF} for each dataset, which were
separated into 50 different train (2/3) and test sets (1/3). We first find the average misclassification rates (MR) and 
number of features used for each of the 50 resamples, eliminating at this step the \texttt{mtry} column. Out of that, we  filter by the resample with the smallest MR. According to Table \ref{table:classification_results}, the Standard RF uses more features, but does not always have the 
lowest MR. As for the regularization, using $g(\mathbf{x}_i) = \Big(\frac{\text{MutInf}(\mathbf{y}, \mathbf{x}_i)}{max_{j=1}^{P}\text{MutInf}( \mathbf{y})} \Big)$ is better for [brain]  and [prostate], while when $g(\mathbf{x}_i) = \text{Boosted}_{RF}$,  the results are good for the 
[adenocarcinoma], [breast 2], [breast 3] and [nci 60], and the \textit{GRRF} is strictly better for the [colon] and [srbct] datasets, considering the MR. The MRs are all the same for the [leukemia] and [lymphona] 
datasets, but the percentage of features is often 
the lowest for our gain penalization method,  especially when 
$g(\mathbf{x}_i) = \Big(\frac{\text{MutInf}(\mathbf{y}, \mathbf{x}_i)}{max_{j=1}^{P}\text{MutInf}( \mathbf{y})} \Big)$. When this happens and such algorithms also have a low or very similar MR to a Standard RF one, we reach an optimal situation, which happened for almost all the datasets. 

The comparison to LASSO  \cite{Tibshirani1991} and varSelRF \cite{DiazUriarte2007}
is presented in the Appendix A 1.2 (only 10 resamples
were done for these models due to their computational heaviness).

\section{Implementation}

The implementation is included as an extension to the \texttt{ranger} package \citep{ranger},
which is originally written in \texttt{C++}
and is currently the fastest tree model implementation 
available for \texttt{R}  \citep{rmanual}.  Furthermore,
the package has a wide variety of model extensions, 
is actively maintained and interfaces with
\texttt{python}. The speed and scalability discussion presented in \citet{ranger}
and its comparison to the \texttt{randomForest}
package \citep{rf_package_paper} is  analogous to the one
about our regularization
implemented in the \texttt{ranger} and the one in the \texttt{RRF}
package, so we do not repeat the same experiments\footnote{\textit{All the code and data used are available in a GitHub repository, now kept hidden for blind review purposes.}}.

\section{Conclusions and next steps}

Feature selection and regularization for tree-based methods is 
a topic of active research. In this work, we have demonstrated that our gain 
penalization generalization,
which combines previous information about the features with a baseline 
penalization $\lambda_0$, produces good results in terms of the 
(number of features) x (prediction error) trade-off. Along with the methodology, we make the implementation available in the fastest Random Forest package for \texttt{R}.

The downside of our approach is the addition of new hyperparameters, and
how to choose them well. Future work involves finding theoretical properties of certain gain penalization approaches, parameter optimization (using e.g. \citet{bayesopt}), and compare our approach to other methods
with a similar context (\citet{johnson2013learning, nan2017adaptive, nan2016pruning}, for example).




\small

\bibliography{references}
\bibliographystyle{neurips_2020}

\newpage 
\appendix

\section{Appendix}

\subsection{Dataset descriptions}

\begin{table}[h]
\centering
\caption{Datasets' specifications. Problematic $p > n$ situation in all cases.}
\label{table:classification}

\begin{tabular}{lcccc}
  \hline
  \hline
Dataset & Ref. & Obs. & Features & Classes \\ 
  \hline
  \\[-8pt]
  adenocarcinoma &  [\citenum{ramaswamy2003molecular}] & 76  & 9869 & 2 \\
  brain          &  [\citenum{pomeroy2002prediction}]  & 42  & 5598 & 5 \\ 
  breast 2       &  [\citenum{van2002gene}]            & 77  & 4870 & 2 \\ 
  breast 3       &  [\citenum{van2002gene}]            & 95  & 4870 & 3 \\ 
  colon          &  [\citenum{alon1999broad}]          & 62  & 2001 & 2 \\ 
  leukemia       &  [\citenum{golub1999molecular}]     & 38  & 3052 & 2 \\ 
  lymphoma       &  [\citenum{alizadeh2000distinct}]   & 62  & 4027 & 3 \\ 
  nci 60         &  [\citenum{ross2000systematic}]     & 61  & 5245 & 8 \\
  prostate       &  [\citenum{singh2002gene}]          & 102 & 6034 & 2 \\ 
  srbct          &  [\citenum{khan2001classification}] & 63  & 2309 & 4 \\ 
   \hline
   \hline
\end{tabular}
\end{table}

\newpage

\subsection{LASSO and varSelRF results}

In Table \ref{table:classification_results}, we present
the classification results added with the LASSO
and varSelRF results. Only 10 resamples were done for these algorithms
due to their computational heaviness. We can see that the LASSO
and varSelRF can select very few variables but at the expense
of the prediction error. Their variable selection
results are more closely comparable to our
regularized model  using $g(\mathbf{x}_i) = \Big(\frac{\text{MutInf}(\mathbf{y}, \mathbf{x}_i)}{max_{j=1}^{P}\text{MutInf}( \mathbf{y})} \Big)$,
but their prediction error is much higher. 

\begin{table}[h]
\setlength\tabcolsep{2.5pt} 
\centering
\caption{Average \% of features used and average misclassification rates with standard deviation added with the LASSO and varSelRF results. The regularized models used far fewer features than a Standard RF. Clearly, our approaches use far fewer features whilst maintaining competitive misclassification performance.}
\label{table:classification_results}

\begin{tabular}{l | c c c c c c}
  \toprule
  \multicolumn{1}{l}{} & \multicolumn{4}{c }{\textbf{Percentage of features used}} \\  
  \hline
 Dataset & Std.RF & GRRF & $\text{Bst}_{RF}$ & Mut.Inf.  & LASSO & varSelRF \\ 
  \hline
  \\[-8pt]

adenocarcinoma & 9.38 & 0.83 & 0.86 & 0.07   & \textbf{0.02} &  0.05  \\ 
brain & 25.06 & 1.44 & 1.46 & \textbf{0.14} & 0.39    & 0.73 \\ 
breast 2 & 24.44 & 1.76 & 1.79 & \textbf{0.14} & 0.21 & 0.34 \\
breast 3 & 38.58 & 1.95 & 2.00 & \textbf{0.28} & 0.67 & 0.28 \\
colon & 34.44 & 2.53 & 2.60 & \textbf{0.44} & 0.46  &   0.94 \\
leukemia & 8.07 & 1.26 & 1.25 & \textbf{0.05} &   0.28   &  0.09 \\
lymphoma & 13.88 & 1.07 & 1.14 & \textbf{0.08} & 0.34 & 0.72 \\
nci & 44.77 & 1.81 & 1.88 & \textbf{0.16} & 1.11   &    0.97 \\
  prostate & 18.22 & 1.46 & 1.40 & \textbf{0.09} & 0.14 &   0.07 \\
  srbct & 29.69 & 2.26 & 2.25 & \textbf{0.3} & 0.68  &  0.99 \\
\hline
\end{tabular}

\vspace{3mm}
\begin{tabular}{l | c c c c c c}
  \hline
  \multicolumn{1}{l}{} & \multicolumn{4}{c }{\textbf{Misclassification rate in the test set}} \\  
  \hline
 Dataset & Std.RF & GRRF & $\text{Bst}_{RF}$ & Mut.Inf.  & LASSO & varSelRF \\ 
  \hline
  \\[-8pt]

adenocarcinoma & 3.45 (0.0) & 11.03 (12) & \textbf{3.4} (0.0) & 10.77 (1.7) &  13.83 (6.4)  & 19.6 (7.7)\\ 
brain & 8.00 (5.6) & 15.38 (14.4) & 15.00 (7) & \textbf{13.33 (10.5)}  & 27.6 (11.7) & 29 (16.2)\\ 
breast 2 & 25.6 (5.4) & 20.7 (2) & \textbf{17.8 (7.2)} & 20.7 (7.7)
& 31.56 (5.19) &  36.7 (9.1)\\ 
breast 3 &  27.6 (6.4) & 30.6 (4) & \textbf{28.1 (3.8)} & 29.3 (1.6)
& 29.7 (4.85) & 33.9 (8.6)\\ 
colon & 4.76 (0.0) & \textbf{5.71 (2.1)} & 6.67 (0.0) & 7.78 (3) 
& 16.7 (8.6) & 23.06 (8.3) \\ 
leukemia &  0.0 (0.0) & \textbf{0.0 (0.0)} & \textbf{0.0 (0.0)} & \textbf{0.0 (0.0)} & 10.07 (9.4)  & 13.2 (12)\\ 
lymphoma &  0.0 (0.0) & \textbf{0.0 (0.0)} & \textbf{0.0 (0.0)} & \textbf{0.0 (0.0)} & 1.48 (4.7)&  5.86 (4.8) \\ 
nci & 30.77 (9.4) & 38.95 (8) & \textbf{37.5 (7.6)} & 43.75 (0.0) &
42.2 (7.5) & 44.7 (12.8)  \\ 
  prostate &  \textbf{0.54 (1.2)} & \textbf{0.54 (1.2)} & 1.08 (1.5) & \textbf{0.54 (1.2)} & 6 (2.8)  & 8.87 (1.9) \\ 
  srbct & 0.0 (0.0) & \textbf{0.0 (0.0)} & 0.91 (2) & 1.74 (3.9) &
  1.33 (2.9)  &  4.5 (3.6) \\ 
\bottomrule
\end{tabular}

\end{table}

\end{document}